\PassOptionsToPackage{unicode}{hyperref}
\PassOptionsToPackage{hyphens}{url}
\documentclass[
]{article}
\usepackage{xcolor}
\usepackage{amsmath,amssymb}
\usepackage{iftex}
\ifPDFTeX
  \usepackage[T1]{fontenc}
  \usepackage[utf8]{inputenc}
  \usepackage{textcomp} % provide euro and other symbols
\else % if luatex or xetex
  \usepackage{unicode-math} % this also loads fontspec
  \defaultfontfeatures{Scale=MatchLowercase}
  \defaultfontfeatures[\rmfamily]{Ligatures=TeX,Scale=1}
\fi
\usepackage{lmodern}
\ifPDFTeX\else
\fi
\IfFileExists{upquote.sty}{\usepackage{upquote}}{}
\IfFileExists{microtype.sty}{% use microtype if available
  \usepackage[]{microtype}
  \UseMicrotypeSet[protrusion]{basicmath} % disable protrusion for tt fonts
}{}
\makeatletter
\@ifundefined{KOMAClassName}{% if non-KOMA class
  \IfFileExists{parskip.sty}{%
    \usepackage{parskip}
  }{% else
    \setlength{\parindent}{0pt}
    \setlength{\parskip}{6pt plus 2pt minus 1pt}}
}{% if KOMA class
  \KOMAoptions{parskip=half}}
\makeatother
\usepackage{longtable,booktabs,array}
\usepackage{calc} % for calculating minipage widths
\usepackage{etoolbox}
\makeatletter
\patchcmd\longtable{\par}{\if@noskipsec\mbox{}\fi\par}{}{}
\makeatother
\IfFileExists{footnotehyper.sty}{\usepackage{footnotehyper}}{\usepackage{footnote}}
\makesavenoteenv{longtable}
\usepackage{graphicx}
\makeatletter
\newsavebox\pandoc@box
\newcommand*\pandocbounded[1]{% scales image to fit in text height/width
  \sbox\pandoc@box{#1}%
  \Gscale@div\@tempa{\textheight}{\dimexpr\ht\pandoc@box+\dp\pandoc@box\relax}%
  \Gscale@div\@tempb{\linewidth}{\wd\pandoc@box}%
  \ifdim\@tempb\p@<\@tempa\p@\let\@tempa\@tempb\fi% select the smaller of both
  \ifdim\@tempa\p@<\p@\scalebox{\@tempa}{\usebox\pandoc@box}%
  \else\usebox{\pandoc@box}%
  \fi%
}
\def\fps@figure{htbp}
\makeatother
\NewDocumentCommand\citeproctext{}{}

\makeatletter
 \let\@cite@ofmt\@firstofone
 \def\@biblabel#1{}
 \def\@cite#1#2{{#1\if@tempswa , #2\fi}}
\makeatother
\newlength{\cslhangindent}
\newlength{\csllabelwidth}
\newenvironment{CSLReferences}[2] % #1 hanging-indent, #2 entry-spacing
 {\begin{list}{}{%
  \setlength{\itemindent}{0pt}
  \setlength{\leftmargin}{0pt}
  \setlength{\parsep}{0pt}
  \ifodd #1
   \setlength{\leftmargin}{\cslhangindent}
   \setlength{\itemindent}{-1\cslhangindent}
  \fi
  \setlength{\itemsep}{#2\baselineskip}}}
 {\end{list}}
\usepackage{calc}

\usepackage{newunicodechar}
\newunicodechar{π}{\ensuremath{\pi}}
\usepackage{bookmark}
\IfFileExists{xurl.sty}{\usepackage{xurl}}{} % add URL line breaks if available
\makeatletter
\@ifundefined{xmpquote}{}{}
\makeatother
\hypersetup{
  pdftitle={Prior-matched evaluation of operational Earth-observation classifiers: a three-number reporting method demonstrated on Sentinel-1 internal-wave detection},
  pdfauthor={João Pinelo},
  hidelinks,
  pdfcreator={LaTeX via pandoc}}

\title{Prior-matched evaluation of operational Earth-observation
classifiers: a three-number reporting method demonstrated on Sentinel-1
internal-wave detection\footnote{This manuscript is being submitted to
  \emph{Geoscientific Model Development}.}}
\author{João Pinelo, João Gonçalves, Arun Shukla, Adriana
Santos-Ferreira\footnote{Atlantic International Research Centre (AIR
  Centre), Azores, Portugal. ORCID --- João Pinelo: 0000-0002-4890-0775;
  João Gonçalves: 0009-0001-4547-1696; Arun Shukla: 0009-0001-5812-1660;
  Adriana Santos-Ferreira: 0000-0002-5704-6021. Correspondence: João
  Pinelo (joao.pinelo@aircentre.org)}}
\date{19 September 2026}

\begin{document}
\maketitle
\begin{abstract}
The Internal Waves Service screens the Sentinel-1 Wave-mode archive for
internal solitary waves, routing detections to experts whose time is the
scarce resource, so precision leads. Its classifier was trained and
reported at a one-to-one class balance, fixed before the operational
rate could be known; that rate has since emerged as heavily skewed, of
the order of one scene in twenty, and a balanced-test score overstates
the precision a validator meets. The incumbent scores 0.794
balanced-test precision and 0.192 in operation. At a fixed recall, prior
correction and calibration are monotone transforms of the score and
cannot move precision, so the mismatch is an evaluation problem dressed
as a training one. We answer it by reporting three numbers,
balanced-test, operational-prior and real post-deployment, whose
contrast is the honest measure. A precision-first, leakage-controlled
cycle improves the classifier lever by lever, each step promoted against
a pre-registered margin: negative variety and the aggregation head lift
precision, capacity pays once and then stops, the training ratio and
calibration cannot, the negatives being as much a result as the gains.
Holding recall at a floor of 0.80 and certifying against a sealed,
single-read lockbox, the promoted model reports 0.927 precision at the
operational prior, and the pre-registered deployment window fields
0.7936 at recall 0.669: it delivers in production what its predecessor
could only report. The prediction held in direction, not in magnitude,
and is reported as promised; with the area under the receiver operating
characteristic curve at 0.933 against the sealed set's 0.993, what
remains is discrimination on the live stream rather than a misplaced
threshold. The window also read the review queue: the prior itself is an
estimate still being discovered, which the next cycle should measure
rather than inherit.
\end{abstract}

\subsection{1. Introduction}\label{introduction}

Internal solitary waves are among the most energetic features of the
ocean, and their surface signature --- alternating bands of increased
and reduced roughness as the wave-induced currents strain the short
surface waves that synthetic-aperture radar (SAR) responds to --- is
recorded in SAR imagery, which makes a satellite SAR archive a natural
instrument for mapping where and how often they occur (e.g., Barintag et
al. 2023). The Internal Waves Service builds that map operationally from
Sentinel-1 Wave-mode data, running a convolutional classifier over the
incoming vignettes and passing its detections to domain experts who
confirm or reject each one. The classifier's purpose is not to be the
final word but to spend the experts' time well: the archive is far
larger than any team could inspect exhaustively (Sect. 2), so the model
exists to concentrate scarce expert attention on the scenes most likely
to carry a wave.

That framing sets the cost of error. A false positive lands a wave-free
scene on a validator's desk and consumes the attention the service is
built to conserve; a missed wave is largely recoverable, because
Wave-mode acquisitions recur over the same locations on the orbit repeat
and the archive is reprocessed, so the same site returns for another
look. Precision therefore leads and recall is held to a floor rather
than maximised --- the operating discipline the whole effort is
organised around.

The difficulty is not the model so much as how its skill is reported.
The classifier was trained and evaluated, by the convention it
inherited, on class-balanced data --- equal numbers of wave and non-wave
scenes --- while the stream it actually runs on carries a positive rate
of only about 0.05. A model that looks well-behaved on a balanced test
set over-fires badly on that skew, and the balanced-test figures give no
warning of it, because they never present the model with the
negatives-to-positives ratio it will meet in production. The gap between
the reported number and the fielded one is not noise or a tuning
oversight; it is a systematic artefact of reporting at the wrong prior,
and it is invisible precisely to the metric most papers quote (Dockès et
al. 2021). Closing that gap --- making the reported figure predict the
operational one --- is the problem this paper takes up, and the method
it proposes is the contribution.

It would be natural to treat a prior mismatch as a training defect and
correct it by retraining at the operational ratio, or by recalibrating
the model's probabilities to the deployment prior (Saerens et al. 2002;
Heiser et al. 2020). Neither helps where it counts. At a fixed recall
--- the operating discipline above --- prior correction and probability
calibration are monotone transformations of the score and cannot move
precision at all; they relocate the threshold, not the curve (Sect.
4.3). The mismatch is therefore an evaluation problem wearing the
costume of a training one, and the response it calls for is an honest
reporting method, not a prior-matched retrain. That inversion is the
paper's organising claim.

The contribution is threefold. First, a prior-matched reporting method:
an operational classifier is characterised by three figures rather than
one --- its balanced-test performance, its performance on a frozen test
set drawn at the operational prior, and its real operational performance
from post-deployment expert adjudication --- and the methodological
content lies in the contrast among them, the balanced figure exposed as
optimistic and the operational-prior figure built to predict the fielded
one (Sect. 3). Second, a worked demonstration: a precision-first,
leakage-controlled development cycle that improves the deployed
classifier lever by lever, each intervention isolated and promoted only
against a pre-registered margin, and each reported for what it did and
did not buy --- including the levers that bought nothing (Sect. 4).
Third, the discipline behind the demonstration: a dataset pinned before
it is split, spatial leakage controlled at the level of the true image
footprint, an operating point fixed on development data, and a sealed
evaluation set read exactly once at a pre-registered threshold --- so
that the honest negatives and the single-read certification are
themselves a reproducibility result (Sects 3, 5). A by-product is itself
a first: the Internal Waves Service is, to our knowledge, the first
operational classifier built to screen the whole Sentinel-1 Wave-mode
archive for internal solitary waves, so the operational positive rate
its adjudicated stream reveals --- roughly one scene in twenty (Sect.
2.3) --- is an initial empirical estimate of that rate, the very prior
that was unmeasurable when the balanced training convention was fixed.

This is a successor to the pipeline-selection study that established the
service's classifier (Pinelo et al. 2026), not a correction of it. That
work chose a deployable network on operational grounds under the
balanced convention then in force, and the choice was sound on the
evidence available. What changed is the ground beneath it: the
operational positive rate became known, more validated data accrued, and
the question shifted from which pipeline to field to how far the fielded
one can be pushed and how its skill should be reported. The present
paper restarts from a smaller deployed network rather than the larger
candidate, and takes it as far as the data allow --- reaching the
conclusion that the binding constraint has moved off the model and back
onto the data (Sect. 4.6).

Section 2 sets out the service and the evaluation problem in operational
terms; Section 3 defines the three-number instrument and the frozen,
leakage-controlled test set behind it; Section 4 documents the
development cycle lever by lever; Section 5 reports the instrument's
output on the promoted model and its sealed certification; and Sections
6 and 7 draw out what transfers to other operational Earth-observation
services and what the next cycle requires.

\subsection{2. The Internal Waves Service and the evaluation
problem}\label{the-internal-waves-service-and-the-evaluation-problem}

\subsubsection{2.1 The service}\label{the-service}

The Internal Waves Service classifies the operational Sentinel-1
Wave-mode archive for internal solitary waves, working from the short
vignettes the mission acquires continuously along-orbit (Santos-Ferreira
et al. 2025; Pinelo et al. 2025). The growing catalogue currently holds
over 1.7 million vignettes, a fraction of the full Sentinel-1 Wave-mode
archive, which we estimate at approximately 17 million vignettes; it
grows through a continuous harvest, so any evaluation set must be frozen
against a moving target (Sect. 3.2). Model detections are not taken at
face value: they enter a review queue where domain experts adjudicate
them, and the queue is organised by a model-confidence band so reviewers
can concentrate where their judgement is most useful. The expert time
that adjudication consumes is the resource the whole effort exists to
conserve.

\subsubsection{2.2 Why precision leads}\label{why-precision-leads}

The two error types do not cost the same. A false positive spends an
expert's attention on a scene that holds no wave; a missed positive is
largely recoverable, because the same location is reimaged on the 12-day
orbit repeat and the archive is periodically reprocessed, so a wave
missed on one pass has further chances to be caught. Precision therefore
leads, and recall is held at a floor of 0.80 rather than pursued as a
target (Sect. 3.3). This is not merely operational convenience: the
asymmetry is what makes a false positive the expensive error, and it
sets the direction of the entire development effort --- lift precision,
hold recall.

That precision leads is itself the current end of an inversion, and the
history matters because it explains the class balance the model
inherited. The service began recall-first. With few confirmed examples
to learn from, the early priority was to harvest positives aggressively
and build a labelled set, and a missed wave then was a training example
lost rather than a recoverable operational miss --- the low default
decision cut was set in that spirit, to leave few real positives
unnoticed and route the boundary cases to human labelling. The training
data were assembled at an even, one-to-one class balance, a ratio fixed
at the outset around the service's competition origins, when the
operational positive rate was not yet known, and that convention was
carried forward unexamined through each successive model. The ground has
since shifted: the operational rate is now known to be heavily skewed
(Sect. 2.3), the positive set is large enough that harvesting is no
longer the constraint, and the cost that now binds is the validator time
each false positive consumes. Hence precision-first, with recall held at
its floor rather than maximised. The stance is the current one, not a
permanent one --- as the two error costs converge, with a larger
positive backlog and a service no longer starved of examples, the
emphasis is expected to normalise toward F1 --- but the recall floor of
0.80 is held throughout.

Recoverability, the property that lets recall sit at a floor at all,
rests on more than the orbit repeat. Each improved model reprocesses the
entire catalogue rather than only scoring new acquisitions, so a
positive missed by one model generation has a further chance to be
caught by the next: the recoverable miss is recoverable across model
iterations, not only across passes of the same model. This is a
deployment property the service was built for. The predecessor work
chose a fast, energy-frugal Julia inference stack precisely so that
reclassifying the whole archive with a new model is a same-day operation
(Pinelo et al. 2026), and reprocessing wholesale --- rather than
layering each model's verdicts over the last --- keeps the published
catalogue labelled consistently under a single model version, which
matters for a citeable, DOI-versioned dataset. The recall floor is
therefore underwritten by an operational guarantee: what one model
misses, its successor re-examines.

\subsubsection{2.3 Starting state, and what a threshold can
buy}\label{starting-state-and-what-a-threshold-can-buy}

The deployed model is a compact 34,465-parameter convolutional network
(SAR\_CNN v2), trained and validated under an even class balance --- the
convention inherited from the service's origins --- and then applied to
a stream whose true positive rate is only about 0.05 --- now measurable
from the service's own adjudicated detections, but unknowable when the
balanced convention was set. That figure is an estimate drawn from those
adjudicated detections rather than a stream census, and it sharpens as
validation reaches further across the score range. The mismatch surfaces
as over-firing. At the default 0.5 decision cut the model attains a
recall of 0.963 but a precision of just 0.192 on the adjudicated set,
flagging 26,472 false positives; and because validators preferentially
review likely positives, even that precision is optimistic against the
live stream rather than pessimistic.

Raising the decision threshold recovers precision without retraining. At
0.80 the false positives fall to 6,193 --- a reduction of 76.6\% --- for
a recall that eases only from 0.963 to 0.818, lifting precision to 0.464
(Fig. 1). This is the first and cheapest lever (Sect. 4.2): genuine,
immediate relief for the validators, but a redistribution of errors
along a fixed curve, not a lift of it --- precision at the recall floor
stops well short of the target. Two facts frame everything that follows:
the remaining gap is a model problem, not a threshold one, and the
operational precision quoted here rests on the validated subset, so the
honest question is not what the balanced test set reports but what the
skewed stream would. The first motivates the development cycle (Sect.
4); the second motivates the evaluation instrument (Sect. 3).

\pandocbounded{\includegraphics[keepaspectratio]{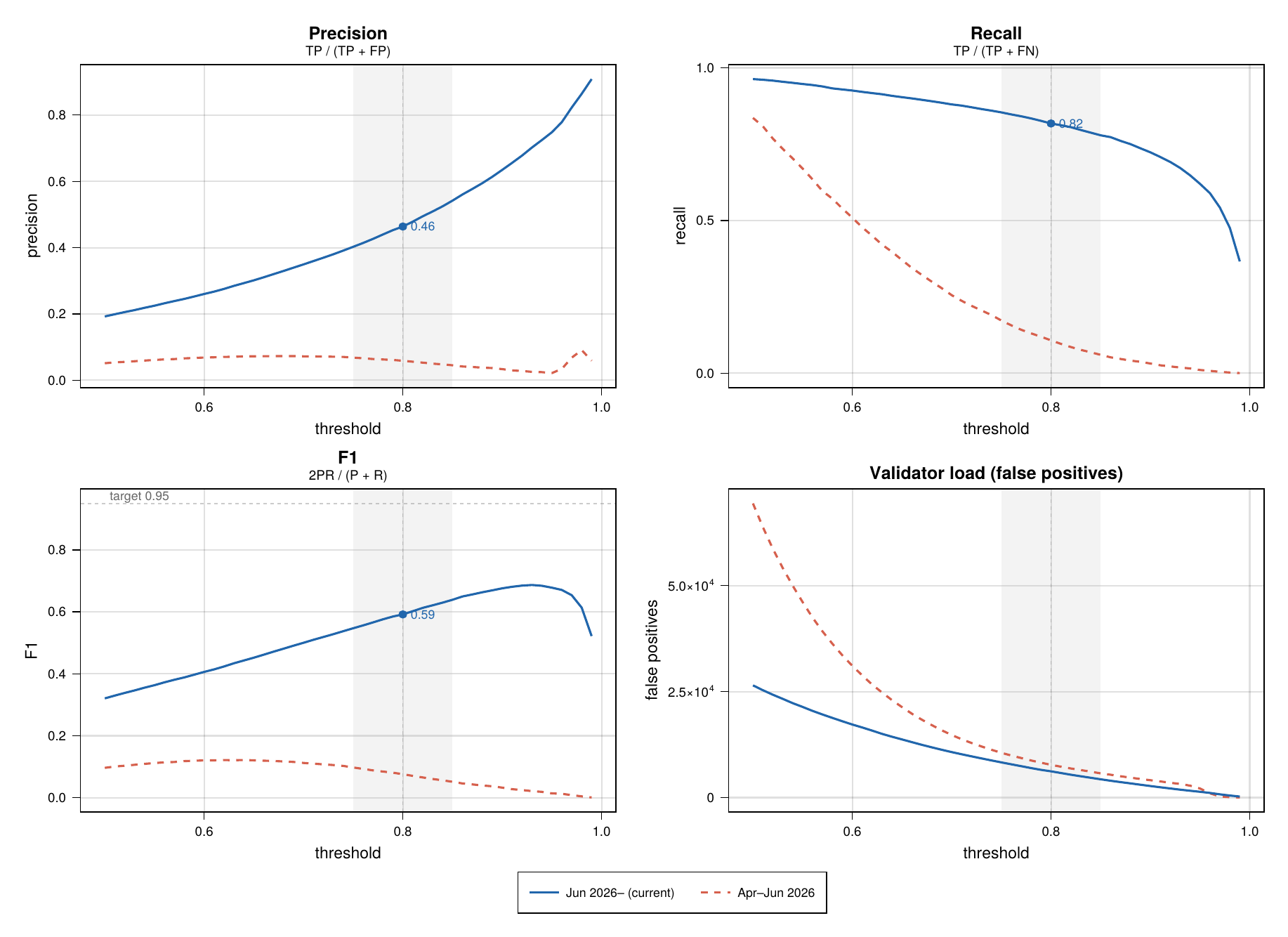}}

\textbf{Figure 1.} Precision, recall, F1 and validator load
(false-positive count) as a function of the decision threshold, for the
two classifiers the service has operated. The current model
(\texttt{20260602-01}, solid) carries the operating-point decision; the
retired April--June model (dashed) is drawn behind it, so the pair
survives greyscale. Each rate panel states its own definition. The
high-value 0.75--0.85 threshold band is shaded and the 0.80 candidate
operating point is marked on the current model; the dashed line in the
F1 panel is the project target of 0.95, which no threshold on either
curve reaches --- the gap a threshold cannot close, and what motivates
the model-side development cycle (Sect. 4). The rates are computed on
the validated subset and so run optimistic against the full stream.

\subsection{3. Methods --- the evaluation
instrument}\label{methods-the-evaluation-instrument}

\subsubsection{3.1 A three-number reporting
axis}\label{a-three-number-reporting-axis}

An operational classifier is characterised here by three performance
figures rather than one, and the methodological claim rests on the
relationship among them, not on any single value.

The first, \emph{balanced-test} performance, is the figure a
conventional pipeline reports: precision, recall and F1 on a held-out
set with equal class proportions (Maxwell et al. 2021a). It is what the
Internal Waves Service pipeline reported for its earlier models, and, as
Sect. 5 shows, it is systematically optimistic once the deployment
stream is skewed (Saito and Rehmsmeier 2015).

The second, \emph{operational-prior} performance, scores the same model,
at the same decision threshold, on a frozen test set drawn at the
operational positive rate (0.05; Sect. 2) rather than at parity (Maxwell
et al. 2021b). This is the honest predictor of fielded behaviour: it
asks what precision a validator would actually face, and it is the
figure the present method proposes as the one to report.

The third, \emph{real-operational} performance, is obtained after
deployment, from expert adjudication of the model's own detections on
the live stream. It is the ground truth the second figure is constructed
to predict. Because it requires the model to run in production through
an adjudication window, it is stated as a forward commitment rather than
approximated by any test-set surrogate, and is reported here from a
post-deployment adjudication window (Sects 5.1, 6).

The instrument is the contrast. For a model trained and validated under
class balance, figure (1) diverges sharply from figures (2) and (3): the
cautionary reading, and the reason balanced-test reporting misleads. For
a model reported under this method, figure (2) is built to track figure
(3): the corrective reading. As a preview of the effect (Sect. 5), the
model promoted in Sect. 4 scores a precision of 0.996 on a balanced set
and 0.927 at the operational prior --- the same model at the same
threshold, the difference being the prior alone. Table 1 (Sect. 5) sets
out the full axis.

\subsubsection{3.2 A frozen, leakage-controlled test
set}\label{a-frozen-leakage-controlled-test-set}

The scarce class sets the terms. Confirmed internal-wave positives
number 6,868 against 126,742 confirmed negatives, so the positive
partition, not the negative, bounds the precision at which any figure
can be resolved; it is sized from the confidence interval required on
precision and recall at the recall floor, not from a round fraction. The
dataset is pinned before the split is drawn (20260621T120557Z), so the
continually growing archive cannot leak into a frozen evaluation.

Spatial leakage is controlled at the level of the \emph{hotspot},
defined by true image-footprint overlap rather than a centroid radius.
Sentinel-1 Wave-mode acquisitions recur over fixed locations on the
12-day orbit repeat, and the surface forcing at a site might evolve
little between passes, so near-identical scenes would otherwise straddle
the split and inflate every figure (Kattenborn et al. 2022; Roberts et
al. 2017). Two split regimes follow from this unit: an
\emph{interpolation} split, which separates acquisitions within a
hotspot (the reads reported here), and a stricter \emph{extrapolation}
split, which holds out whole hotspots and region cells to probe
generalisation to unseen sites (Valavi et al. 2019) (a design
provision). The data are partitioned into train, development, and a
sealed lockbox, with the test prior fixed at 0.05 by construction;
composition is 4,466/81,104 (train), 865/16,435 (dev) and 1,537/29,203
(lockbox), positives/negatives (Fig. 2).

The lockbox is the central pre-deployment estimate of operational
precision, and its credibility rests on how it is handled rather than on
its size. It is sealed off-site with a manifest and a per-file checksum,
and read exactly once, at a threshold fixed in advance on the
development set. Every read is pre-registered: the scoring script and
its test are committed before any number is seen, in a two-commit
structure for the consent-gated read. The expected ordering is train
\textgreater= dev \textgreater= lockbox \textgreater= production, the
gap small and downward; a lockbox figure that slightly exceeds dev is
treated as same-prior sampling on a larger independent draw, not as
improvement and not as leakage (Sect. 5).

\pandocbounded{\includegraphics[keepaspectratio]{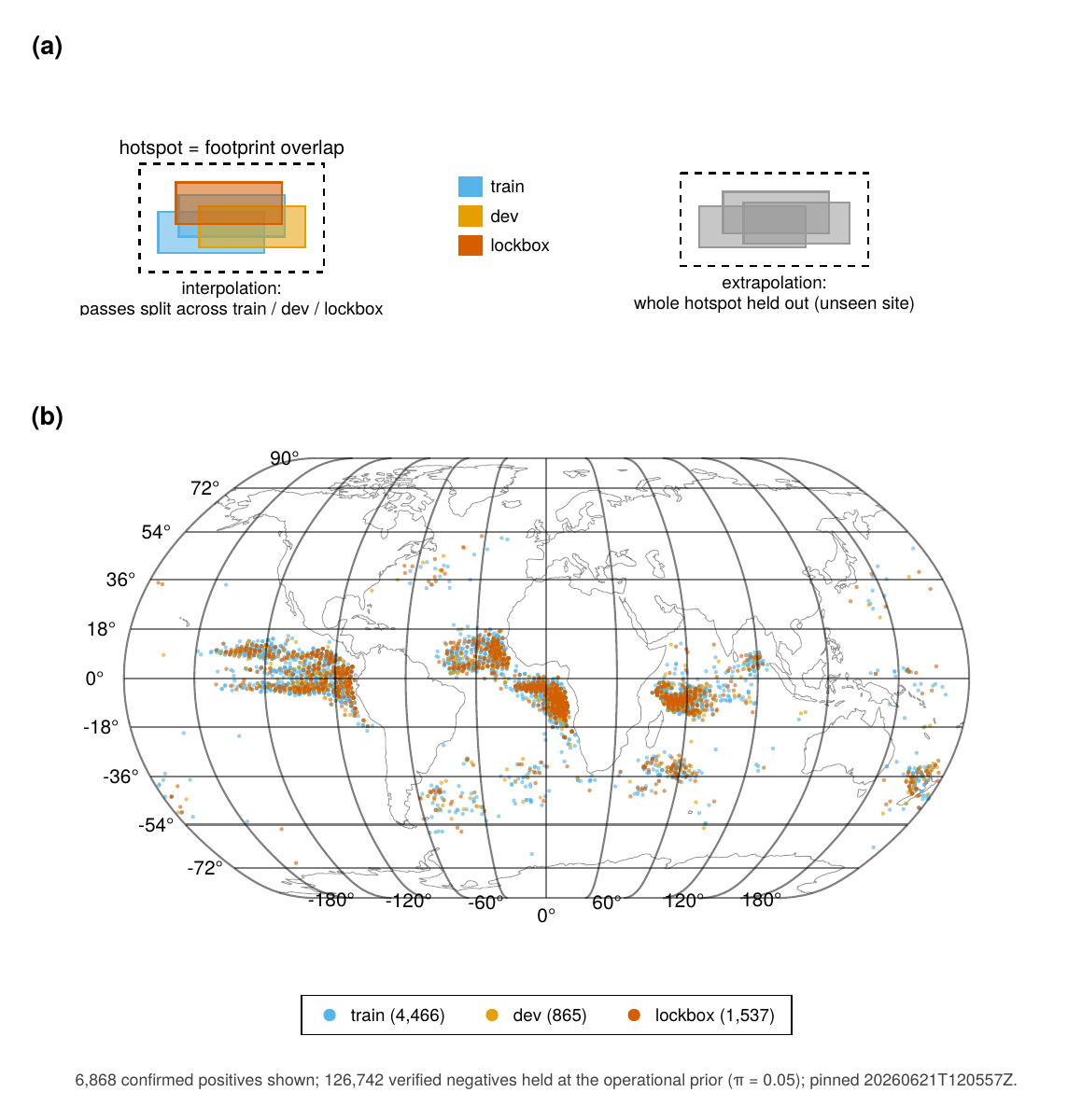}}

\textbf{Figure 2.} The leakage-controlled split. \textbf{(a)} The
leakage unit is the \emph{hotspot} --- a set of Sentinel-1 Wave-mode
acquisitions whose image footprints overlap at a recurrence site. In the
interpolation split used for the reads reported here, a hotspot's passes
are shared across train, development and lockbox; the stricter
extrapolation split (a design provision) instead holds a whole hotspot
out as an unseen site. \textbf{(b)} The 6,868 confirmed positives on a
Natural Earth projection, coloured by split, showing the recurrence-site
clustering the hotspot unit exists to guard against; per-split positive
counts are given in the legend. The 126,742 confirmed negatives, held at
the operational prior (π = 0.05), are a count rather than a geography
and are not mapped. The dataset was pinned (20260621T120557Z) before the
split was drawn, so the continually growing archive cannot leak into it.

\subsubsection{3.3 Operating point and recall
floor}\label{operating-point-and-recall-floor}

The cost of the two error types is asymmetric. A false positive lands on
a validator's desk and consumes expert time; a stochastic miss is
recovered on the next reprocessing pass. Precision therefore leads, and
recall is held at a floor of 0.80 rather than treated as a target:
reprocessing recovers per-pass stochastic misses but not systematic
blind spots, so the floor guards against trading away recoverable recall
while precision is driven up.

The operating point is selected on the development set from a band of
thresholds spanning the recall floor, then projected onto the frozen
lockbox scores; the sealed set never selects its own operating point.
The chosen threshold is 0.935 for the model promoted in Sect. 4.

\subsection{4. Methods --- the development
cycle}\label{methods-the-development-cycle}

\subsubsection{4.1 A moving binding
constraint}\label{a-moving-binding-constraint}

The improvement proceeded as a sequence of single-lever interventions,
each isolated so its contribution could be credited on its own, and each
carried forward only if it beat the incumbent by more than a
pre-registered promotion margin (0.021, the seed-paired 95\% confidence
half-width). The organising observation is that the binding constraint
moved: the lever that paid at one stage was inert at the next, and
reporting where it moved --- rather than only the final number --- is
the point of the demonstration. All development figures below are
precision at the recall floor (0.80), scored at the operational prior on
the frozen development split, across five seeds.

\subsubsection{4.2 Lever 1 --- the operating
threshold}\label{lever-1-the-operating-threshold}

The first and cheapest lever moves the operating point along the
existing model's precision/recall curve, with no retraining, and is
reversible on the next reprocessing pass. It was taken in the
predecessor work (Sect. 2) and delivered most of the immediate
validator-facing relief, but the current curve tops out well short of
the target: a threshold move redistributes errors, it does not lift the
curve. This bounded what remained and set the agenda for the model-side
levers that follow.

\subsubsection{4.3 Lever 2 --- the data}\label{lever-2-the-data}

The dataset was the first model-side lever, and its two effects were
separated rather than moved together: the \emph{class ratio} the model
trains under, and the \emph{negative variety} it sees against the
roughly 95\% of negatives the balanced trainer discards (Kang et al.
2020; Branco et al. 2016). In a factorial over both, negative variety
carried the precision --- widening the negatives lifted operational
precision by 0.045 at the recall floor, a difference that clears zero
--- while shifting the training ratio toward the operational prior did
not, and if anything cost precision (-0.026 for the ratio contrast at
wide variety). The best data configuration, wide variety at a balanced
ratio, reached 0.832 precision at the floor.

This is the first negative result, and it is structural rather than
incidental: prior correction and probability calibration are monotone
transforms of the score (Saerens et al. 2002; Heiser et al. 2020), so at
a recall-pinned operating point they cannot move precision at all ---
they relocate the threshold, not the curve. A separate calibration arm
confirmed the inertness by construction. The prior-matched
\emph{training} the mismatch might seem to call for is therefore the
wrong response to it; the mismatch is an \emph{evaluation} problem,
which is the paper's point. That inertness concerns the reported figure
alone: because these transforms preserve the ranking, they cannot move
the curve, and so cannot improve the precision this method reports ---
but they remain the appropriate instrument for placing the operating
point once the prior shifts, a distinct use returned to under the
out-of-time decay of Sect. 5.3 and Sect. 6.

\subsubsection{4.4 Lever 3 --- capacity}\label{lever-3-capacity}

With the data lever spent at 0.832, capacity was tested by a uniform
width ladder over the deployed three-block shape, architecture the sole
variable. Doubling width promoted --- 0.868 precision at the floor,
0.036 over the data-lever endpoint --- lifting the parameter count from
34,465 to 131,329. But the ladder then flattened: tripling width
(290,657 parameters, near the scale of the predecessor paper's headline
net at 283,329 (Pinelo et al. 2026)) did not clear the promotion margin
at the floor (-0.006, within margin) and quadrupling it (512,449
parameters) likewise (0.006). Beyond the first doubling, added width did
not clear the margin at five seeds --- an interval consistent with a
small residual gain as well as with none, so this is a within-margin
negative rather than a demonstrated ceiling.

The ladder was stopped by a pre-registered two-trigger rule ---
diminishing dev gain within the promotion margin, and the onset of a
train-dev gap --- rather than at a fixed large network. The choice not
to overstretch is a scientific one, not only a regularisation one: a
model overfit to the known recurrence sites stops generalising to the
unseen hotspots the service exists to find, which is why the split holds
sites out (Sect. 3.2) and why the endpoint is matched to the data rather
than maximised.

\subsubsection{4.5 Lever 4 --- the aggregation
head}\label{lever-4-the-aggregation-head}

Capacity being spent, the next lever was the aggregation mechanism ---
the operation that collapses the spatial feature map to a decision,
swapped in isolation on the promoted width network. Replacing
global-average pooling with generalised-mean (GeM) pooling promoted
where added capacity had not: 0.924 precision at the floor, 0.056 over
width. Two attention-pooling heads, with more parameters, did not lift
at all --- again isolating the mechanism, not the capacity, as the
active ingredient. The learned pooling exponent settled well above unity
(mean 4.9, initialised at 3.0), meaning the head weights the peak
spatial responses over the mean.

The mechanism has a concrete target. A recurring geophysical look-alike
--- a banded, quasi-periodic surface signature that mimics the
along-crest periodicity of an internal-wave packet --- forms a small,
persistent tail of the false-positive population, clustered at a handful
of recurring sites. Average pooling washes the banding into a single
mean and fires; a pooling operation that preserves peak spatial
structure can tell it apart. On the cross-seed false-positive core of 40
scenes, the width network fired on 35.8 on average and GeM on 21.2 ---
the mechanism suppressing roughly a third of the core. The residue of 12
is the same banded class rather than a geographic pocket (not
polar-enriched, p \textasciitilde{} 0.58): what remains is a
discrimination limit, not a location (Fig. 3).

\pandocbounded{\includegraphics[keepaspectratio]{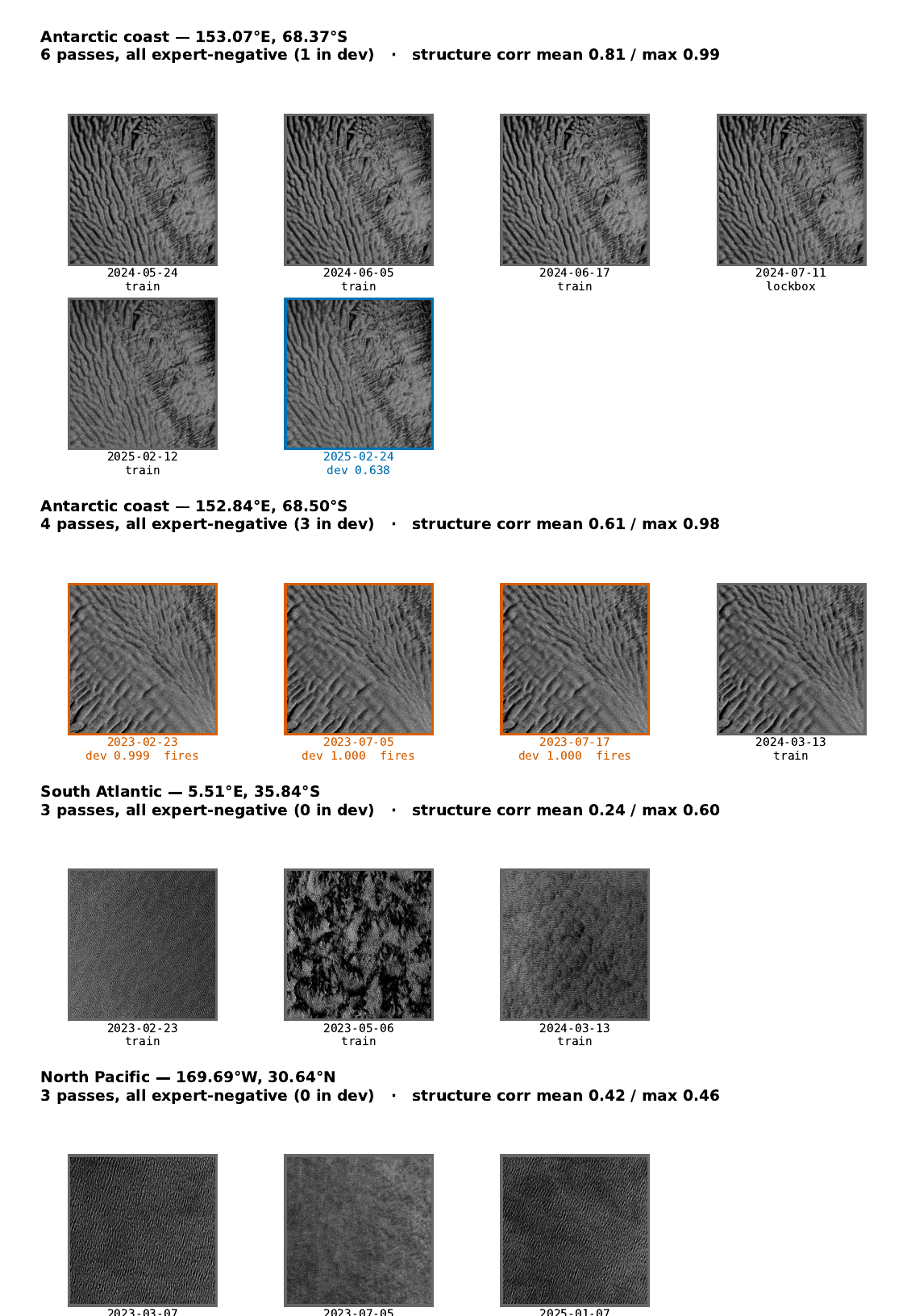}}

\textbf{Figure 3.} The banded geophysical confuser. Sixteen validated
Sentinel-1 WV vignettes from a single acquisition slot (WV slot 025)
that fall at four locations once clustered by centre position --- two
Antarctic-coast footprints, one South Atlantic and one North Pacific
site (each site's clustered position, pass count and per-site
structure-similarity are printed in its header). Every pass at every
site carries an expert-negative verdict. Model behaviour is observable
only on the passes that fall in the development split, and it splits by
site: the three development passes at the 152.84°E Antarctic footprint
fire, all scoring at near-saturation above the recall-0.80 development
threshold, whereas the single development pass at the adjacent 153.07°E
site does not fire; the remaining passes are training or lockbox and
carry no development score (chip borders: vermillion fires, blue
development-no-fire, grey non-development). The high within-site
structure correlation at the two Antarctic footprints --- the
speckle-suppressed 32×32 area-average measure of the data-integrity
audit, near zero between different sites --- establishes that these are
the same banded surface signature recurring on the orbit repeat rather
than duplicated scenes. That signature is a persistent, quasi-periodic
surface texture, consistent with high-latitude sea-ice or wind-banding,
that mimics the along-crest periodicity of an internal-wave packet, and
it names the discrimination target for the aggregation-head work; gem's
suppression of it is reported in the accompanying text, not shown in
this figure. These scenes are drawn from the verified pool, so they are
genuine negatives on which the model is misled by real geophysics, not
by mislabelling.

\subsubsection{4.6 Verdict: where the constraint
moved}\label{verdict-where-the-constraint-moved}

The improved model clears the ranking target it was optimised against
--- 0.924 precision at the recall floor. Its F1 at the floor is 0.860,
short of the 0.90 interim bar and well short of the 0.95 destination. A
precision-ceiling read found no hard architectural wall --- F1
\textgreater= 0.95 is reachable in principle once the reducible
false-positive mass is removed --- but against the full false-positive
mass the best attainable F1 is 0.878, held down by an
irreducible-for-now residue of roughly 57.6 false positives per seed.
Symmetrically, a systematic false-negative core of 34 scenes sits flat
across the recall band, so the operating point is a recoverable-recall
dial above a fixed floor rather than a trade against systematic blind
spots.

Read together, further capacity gains have fallen within the promotion
margin and the aggregation lever is spent, yet there is no ceiling ---
the binding constraint has moved off architecture and onto the size of
the reducible false-positive mass, which is to say onto more validated
data. That is where the next cycle goes, and it is the empirical basis
for restarting from data rather than from a larger network.

\pandocbounded{\includegraphics[keepaspectratio]{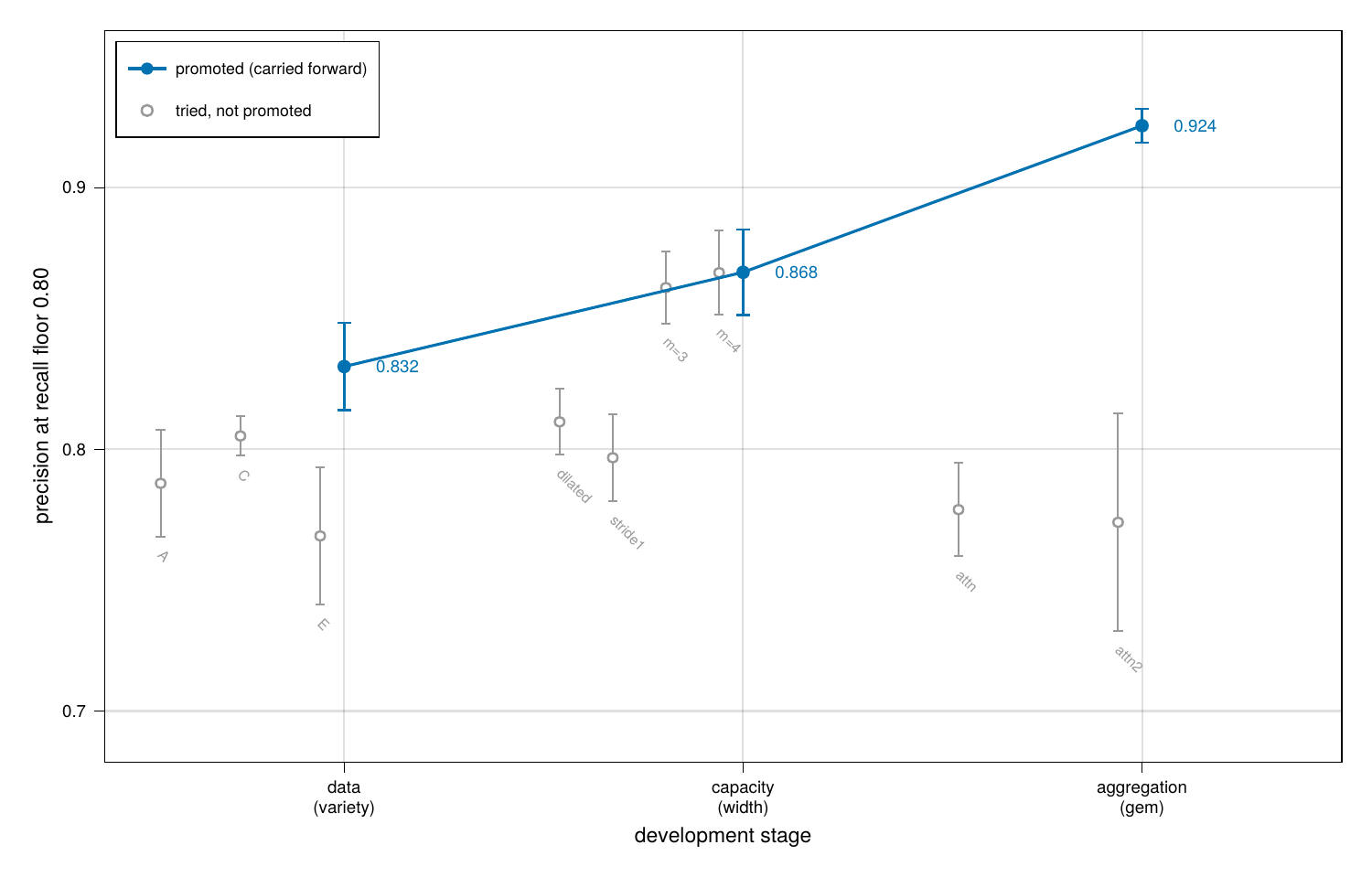}}

\textbf{Figure 4.} The development ladder: precision at the 0.80 recall
floor, scored at the operational prior across five seeds, as the retrain
cycle advanced through its three model-side levers. Each stage's
promoted model (filled, connected) becomes the next stage's baseline, so
the blue spine is the chosen model as it developed --- widening negative
variety (0.832), then doubling width (0.868), then the generalised-mean
(GeM) pooling head (0.924). The rungs tried and not carried forward sit
muted beneath each stage and are named: the discarded data cells at
stage one, the parameter-neutral and higher-capacity architecture
variants at stage two (including the width multiples that did not clear
the margin), and the two attention-pooling heads at stage three. Error
bars are the iteration-1 factorial interval and, for iterations 2 and 3,
the five-seed t-interval; a rung was promoted only on beating the
incumbent by more than the pre-registered margin (0.021, the seed-paired
95\% half-width). Sect. 4.

\subsection{5. Results}\label{results}

The development-cycle results are reported lever by lever in Sect. 4 and
drawn together in Fig. 4; they are not restated here. This section
reports the instrument's output on the promoted model: the three-number
axis (Sect. 5.1), the sealed-lockbox certification behind its
operational-prior cell (Sect. 5.2), an out-of-time robustness check on
that cell (Sect. 5.3), and the operating point set against the
systematic-miss floor (Sect. 5.4).

\subsubsection{5.1 The three-number axis}\label{the-three-number-axis}

Table 1 sets out the three-number axis for the incumbent operational
model and for the model promoted in Sect. 4. The methodological content
is in the contrasts between the cells, not in any single value. Of the
six cells, five are filled from landed reads and one is left n/a by
design (the incumbent's operational-prior cell; see below). The
cautionary contrast (the incumbent's balanced-test figure against its
real operational figure) and the prior-effect contrast (gem's balanced
against its operational-prior figure) can both be read now; the
corrective contrast (gem's operational-prior figure against its real
operational figure) is what the deployment window has now closed.

For the incumbent --- the deployed 34,465-parameter network trained
under class balance --- the cautionary contrast is between its
balanced-test figure and its real operational figure, and both are now
in hand. Its balanced-test performance is precision 0.794, recall 0.964
and F1 0.871 (area under the ROC curve 0.961), computed as the parity
projection of the deployed 0.5 cut over the full verified pool at
complete coverage --- the same operation that produces gem's balanced
cell, so the two balanced figures are comparable by construction rather
than through two different resampling schemes. Its real operational
precision, from validator adjudication at the same 0.5 cut, is 0.192 at
recall 0.963, on the adjudicated subset and so optimistic against the
full stream (Sect. 2). The divergence is the cautionary reading in a
single pair of numbers: a balanced-test precision of 0.794 against a
fielded precision of 0.192 at the identical threshold, the balanced
figure standing far above operational skill because the reporting prior,
not the model, has changed. Two caveats travel with the balanced cell
and neither softens the contrast: the verified pool is review-selected
rather than a stream sample, and the incumbent's June training
membership is unrecoverable --- it is the 34,465-parameter network's own
evaluation, not the predecessor paper's headline figure, which belongs
to a different and larger net. The operational-prior cell is left
unfilled for the incumbent by design: scoring the old model on the
frozen operational-prior set would add nothing its real operational
number does not already show, and the cautionary half rests on the
balanced-versus-real gap, not on an intermediate estimate.

For gem the corrective contrasts are both in hand. At the operational
prior its precision is 0.927, recall 0.827 and F1 0.874 (the sealed
lockbox; Sect. 5.2), and this is the figure built to track the real
operational number the deployment window will produce (Sect. 6). The
same model, at the same threshold, scored on a balanced set reads
precision 0.996, recall 0.827 and F1 0.904 --- the recall unchanged,
because it is a property of the positives alone, and only the precision
moving with the prior. The consequence is the paper's thesis in a single
row: gem's balanced F1 of 0.904 clears the 0.90 interim bar, while its
operational-prior F1 of 0.874 does not --- the balanced figure crossing
a bar the honest figure misses, on identical bytes at an identical
threshold, the difference being the reporting prior and nothing else.

gem's real operational cell is the one number the method cannot borrow
from a test set: it creates its own ground truth prospectively, by
adjudicating gem's detections once gem is the production model. That
window has now run. From 2026-07-22T14:00:00+00 to
2026-08-28T08:45:48+00, over the 41,906 verified images adjudicated
inside it (7,812 positive), gem's real operational precision is 0.7936
at recall 0.669 and F1 0.726, with an area under the ROC curve of 0.933.
The read carries the same discipline as the lockbox: the window's
opening boundary, the generator and its test were committed before any
adjudicated figure was computed, and the closing boundary was fixed as a
rule evaluated at run time rather than as a date chosen once numbers
were visible. Of the 2,587 positives missed at the deployed cut, 2,580
score between 0.01 and the threshold, so the misses are ambiguous-score
scenes rather than confident rejections. The figures are computed on the
adjudicated pool and are not weighted up to the archive: selection
within a review band is deterministic, so the adjudicated images are not
a probability sample of the stream and extrapolation from them would be
unsound. The out-of-time read of Sect. 5.3 remains a robustness check on
the operational-prior cell, not a stand-in for this number. Table 1
therefore stands with every cell filled from a landed read.

\textbf{Table 1.} The three-number reporting axis, precision / recall /
F1 per cell, scored at the dev-fixed threshold (0.935) and, for the
operational-prior and balanced cells, at the operational prior (0.05)
and at parity respectively. The incumbent real-operational figure is
measured on the validated subset and is optimistic against the full
stream (Sect. 2). gem's real-operational figure is the post-deployment
adjudicated read over the window 2026-07-22T14:00:00+00 to
2026-08-28T08:45:48+00, on 41,906 verified images (7,812 positive), and
like the incumbent's is measured on the adjudicated pool rather than a
stream sample. It is reported to four decimals because it coincides with
the incumbent's balanced-test precision to three; the two are unrelated
quantities.

{\def\LTcaptype{none} % do not increment counter
\begin{longtable}[]{@{}
  >{\raggedright\arraybackslash}p{(\linewidth - 6\tabcolsep) * \real{0.2500}}
  >{\raggedright\arraybackslash}p{(\linewidth - 6\tabcolsep) * \real{0.2500}}
  >{\raggedright\arraybackslash}p{(\linewidth - 6\tabcolsep) * \real{0.2500}}
  >{\raggedright\arraybackslash}p{(\linewidth - 6\tabcolsep) * \real{0.2500}}@{}}
\toprule\noalign{}
\begin{minipage}[b]{\linewidth}\raggedright
Model
\end{minipage} & \begin{minipage}[b]{\linewidth}\raggedright
(1) Balanced-test
\end{minipage} & \begin{minipage}[b]{\linewidth}\raggedright
(2) Operational-prior
\end{minipage} & \begin{minipage}[b]{\linewidth}\raggedright
(3) Real operational
\end{minipage} \\
\midrule\noalign{}
\endhead
\bottomrule\noalign{}
\endlastfoot
Incumbent (\texttt{20260602-01}, 34,465 params) & 0.794 / 0.964 / 0.871
& --- (n/a; see text) & 0.192 / 0.963 / 0.321 \\
gem (promoted) & 0.996 / 0.827 / 0.904 & 0.927 / 0.827 / 0.874 & 0.7936
/ 0.669 / 0.726 \\
\end{longtable}
}

\subsubsection{5.2 Lockbox certification}\label{lockbox-certification}

The operational-prior cell rests on a single, sealed read, and its
credibility is a matter of how that read was conducted (Sect. 3.2). The
lockbox holds 30,740 scenes (1,537 positive, 29,203 negative) drawn at
the operational prior and pinned before the split (20260621T120557Z); it
was read exactly once, at the threshold fixed in advance on the
development set (0.935), from the same frozen bytes that will reach
production --- the artefact scored here is byte-for-byte the artefact
that deploys (Sects 3.2, 6).

gem's discrimination survives the move from development to the sealed
set intact: the area under the ROC curve is 0.993 on the lockbox against
0.994 on development, so there is no optimism penalty in ranking. At the
fixed threshold the confusion matrix is 1,271 true and 100 false
positives against 29,103 true and 266 false negatives --- a
false-positive rate of 0.0034 on the sealed negatives, which is the
validator-facing quantity: 100 scenes wrongly flagged against 29,103
true negatives.

One feature of the read is unexpected: the lockbox precision slightly
exceeds the development precision. The lockbox's own recall-0.80
operating point reaches precision 0.939, and at the transferred
development threshold the recall lands at 0.827, above the 0.80 floor.
This is neither improvement nor leakage. The expected ordering is train
\textgreater= dev \textgreater= lockbox \textgreater= production, the
gap small and downward; a lockbox figure a little above dev is
same-prior sampling on a larger independent draw, the ordinary variation
between two finite samples of one distribution, and the split's
construction (Sect. 3.2) forecloses the leakage reading (Fig. 5).

\pandocbounded{\includegraphics[keepaspectratio]{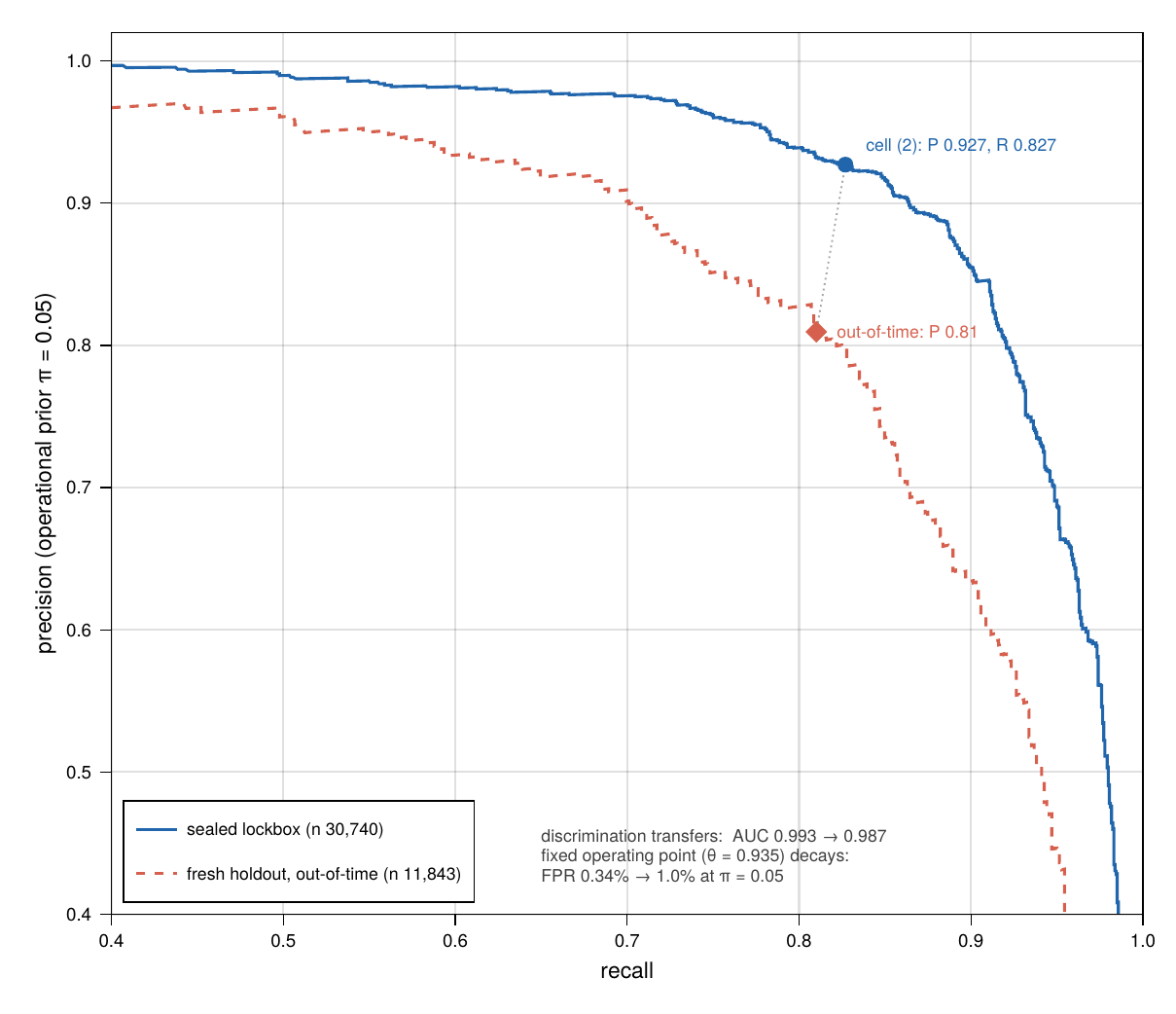}}

\textbf{Figure 5.} Lockbox certification and its out-of-time robustness,
as precision--recall at the operational prior (π = 0.05). The
sealed-lockbox curve (solid, n = 30,740) is the honest pre-deployment
predictor of operational precision; the fresh-holdout curve (dashed, n =
11,843) scores the same deploy bytes at the same fixed threshold (0.935)
on verified scenes whose last verdict post-dates the split-lock pin
(20260621T120557Z) by about ten days, anti-joined off every split
member. The two findings are separate. Discrimination transfers --- the
curves nearly coincide and the area under the ROC curve moves only from
0.993 to 0.987. The fixed operating point (marked on each curve, joined
by the dotted line) does not: precision falls from 0.927 on the sealed
set to 0.810 out-of-time, and the false-positive rate rises roughly
threefold (0.0034 to 0.010) at the same threshold. Reported exploratory
and pre-registered: a single out-of-time window on the review-selected
pool, not a trend (Sect. 5.3).

\subsubsection{5.3 Out-of-time robustness
check}\label{out-of-time-robustness-check}

An independent, out-of-time check on the operational-prior cell has been
run, and is reported here as an exploratory, pre-registered result. It
scores the same deployed bytes at the same fixed threshold (0.935) on
11,843 verified scenes whose last verdict post-dates the split freeze
(20260621T120557Z), anti-joined against every split member so none could
have leaked into training or into the split's selection. The read
separates into two findings. Discrimination transferred intact: the area
under the ROC curve is 0.987 out-of-time against 0.993 on the sealed
lockbox. The fixed operating point did not: at the operational prior the
precision fell from the lockbox's 0.927 to 0.810, and the false-positive
rate rose from 0.0034 to 0.010 at the same threshold, over a window of
only about ten days past the freeze. This is not a composition artefact
--- the drop holds across a prior-sensitivity band (0.771 to 0.837) and
the average precision fell in step (0.890 against the lockbox's 0.944),
so the precision--recall curve genuinely moved rather than the scoring
prior.

Two tiers of caveat of the result. The full read is a real estimate but
a single out-of-time snapshot on the review-selected verified pool ---
one window, not a trend. The footprint-disjoint new-site subset is
underpowered on positives (736 scenes, 20 positive): its AUC 0.986
suggests the model is not blind to unseen geography, but its precision
and recall are not reportable at that sample size.

The finding strengthens rather than dents the thesis. Discrimination is
stable out-of-time while a fixed development-set operating point is not,
and a balanced-test evaluation would have surfaced none of it. That a
falsifiable, pre-registered check caught real out-of-time decay is the
case for prior-matched, out-of-time evaluation; it also sets a concrete
deployment requirement --- that the operating point be validated against
a current catalogue slice rather than inherited from the development set
(Sect. 6). Because the check was pre-registered, with the scoring script
and its test committed before any number was seen, it stands
report-regardless (Fig. 5).

\subsubsection{5.4 The operating point and the systematic-miss
floor}\label{the-operating-point-and-the-systematic-miss-floor}

Applying the fixed threshold places the lockbox at recall 0.827, above
the 0.80 floor, and the recall it forgoes above the floor is the
recoverable kind rather than a systematic blind spot. The systematic
false-negative core --- the positives missed across model families,
cross-architecture and cross-recipe --- numbers 34 scenes and sits flat
across the recall band from 0.70 to 0.85, so moving the operating point
within the band trades stochastic misses, recovered on the next
reprocessing pass, not systematic ones. At recall 0.80 on the deploy
seed the positives forgone number 173, of which the systematic core is
the fixed floor and the remainder the recoverable tail. That floor
corroborates, on a different model and population, the deployed
network's own sub-threshold miss rate of 0.038 (Sect. 4; corroborating,
not load-bearing). The operating point is therefore a recoverable-recall
dial above a fixed floor, which is what the recall-floor discipline
requires (Fig. 6).

\pandocbounded{\includegraphics[keepaspectratio]{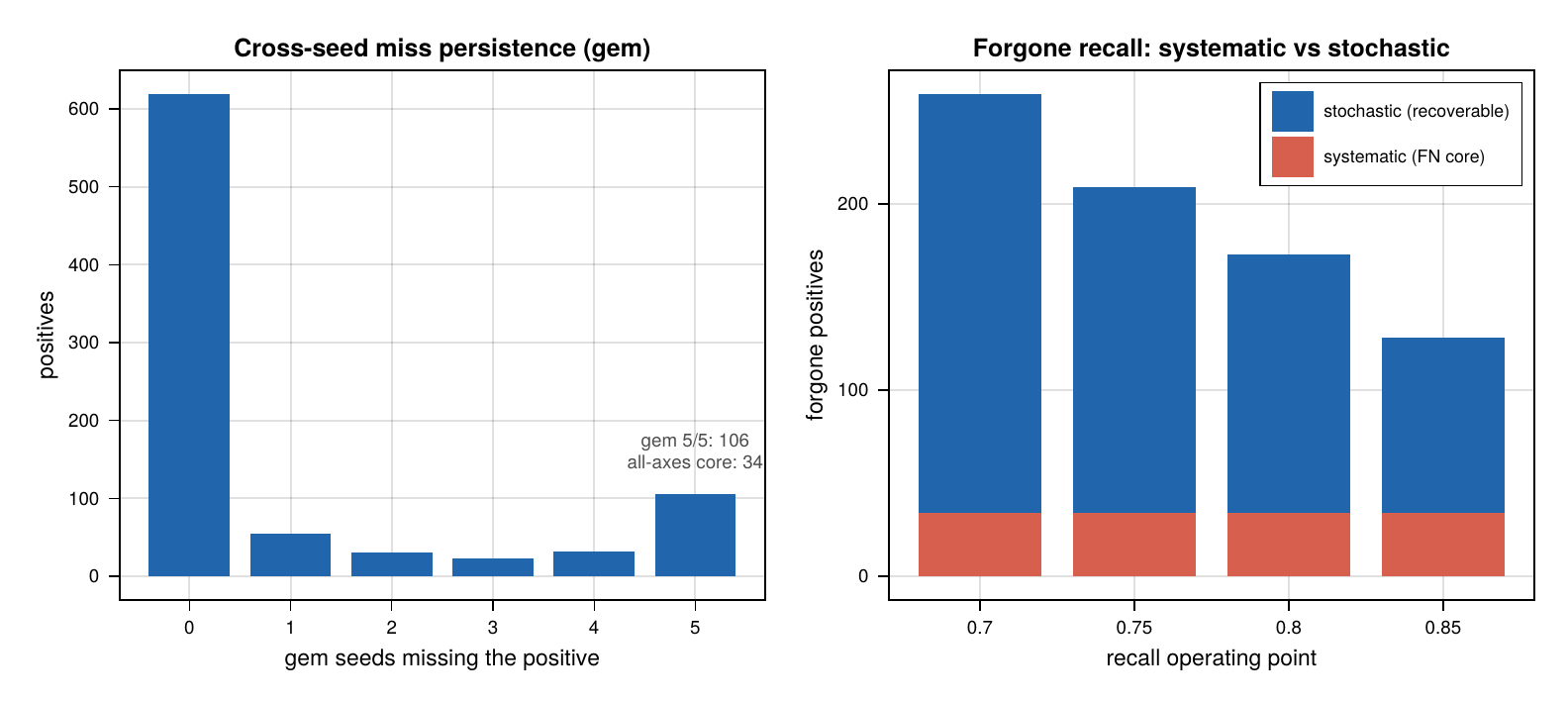}}

\textbf{Figure 6.} The systematic false-negative core, on the
development positives. \textbf{(a)} For each confirmed positive, the
number of the promoted model's five seeds that miss it: most are missed
by no seed, while a persistent tail (106 positives) is missed by all
five. The subset missed across every model family, recipe and seed ---
the systematic core --- is 34 positives. \textbf{(b)} Positives forgone
at each candidate recall operating point, with the systematic core (red)
drawn inside the forgone total (blue) so the stochastic, recoverable
remainder reads off directly. The core is flat across the band, so
moving the operating point within it trades recoverable misses, not
systematic blind spots. Because the development positives are drawn from
the verified pool, the core is a floor: it under-states the
systematic-miss rate on the operational stream.

\subsection{6. Discussion}\label{discussion}

The instrument gives two readings, and the paper's claim is the pair.
Read against the incumbent, it is cautionary: a network trained and
evaluated under class balance carries a balanced-test figure that stands
far above the precision a validator actually meets on the skewed stream,
and the gap is not a defect of the model but an artefact of reporting at
the wrong prior (Sect. 5.1). Read against the promoted model, it is
corrective: the operational-prior figure is constructed to answer the
question the balanced figure evades --- what precision the fielded model
will deliver --- and it is the number the method proposes as the one to
report. The contribution is not either figure alone but the discipline
of carrying both, so that the distance between them is visible rather
than hidden.

The corrective half rests on a prediction this paper pre-registered, and
that prediction is now closed. It did not hold. The operational-prior
cell reads 0.927 precision at recall 0.827; the deployment window
returned 0.7936 at 0.669. Three features of that gap separate. It runs
in the direction the method expects --- the ordering train \textgreater=
dev \textgreater= lockbox \textgreater= production holds (Sect. 3.2) ---
but the margin is wider than the small, downward gap that section
anticipates. It is not explained by the reporting prior: the adjudicated
window pool carries 7,812 positives among 41,906 images, a far higher
positive rate than the lockbox's π = 0.05, and a higher prior raises
precision at a fixed threshold rather than lowering it, so composition
runs against the shortfall rather than accounting for it. And it is not
merely the operating point sitting wrong: the area under the ROC curve
falls from 0.993 on the sealed set to 0.933 operationally, so
discrimination itself is lower on the live stream --- and prior
correction and calibration, which relocate a threshold without moving a
curve (Sect. 4.3), cannot recover a curve that has moved.

Set against that shortfall is the improvement in the number that counts.
The incumbent scored 0.794 balanced-test precision and 0.192 in real
operation; gem scores 0.996 balanced-test and 0.7936 in real operation.
That gem's fielded precision matches the incumbent's balanced-test
figure to three decimals is arithmetic accident, but it puts the case in
one line: the promoted model delivers in production what its predecessor
reported only under a prior it never met. Reporting the shortfall beside
the gain is what the method obliges --- a pre-registered number is
reported whichever way it falls, and a prediction that closes
imperfectly is a finding about the instrument's limits rather than
grounds for withholding it. What the present paper can show is that the
operational-prior figure behaves as a predictor should under the one
independent test available before deployment: on out-of-time data, past
the freeze and disjoint from the split, the model's discrimination held
(Sect. 5.3). The same check also delivered the sharpest caution in the
paper --- the fixed operating point did not transfer unattended, its
precision decaying materially over a short out-of-time window while the
ranking held. That the operational-prior figure survives out-of-time as
a ranking while the threshold placed on it drifts is the distinction the
method draws: the number is a stable ranking, and the condition under
which its operating point predicts is stated rather than assumed.

The window read the review queue as well as the model, and that is a
result of the method rather than a qualification on it. The adjudicated
pool's composition follows review policy: the queue surfaces the scenes
the model scores most confidently positive, so the pool is enriched in
positives, and the window pool's 7,812 positives among 41,906 images sit
far above the 0.05 the frozen test set is composed at. The operational
prior is therefore an output of this method rather than an input to it.
A service adopting the three-number axis bootstraps its prior from a
queue with the same property, and its estimate of that prior sharpens as
adjudication reaches across the score range rather than concentrating on
the detections it is confirming. Knowing that the figure is still being
discovered is part of what the instrument delivers. Estimating a
prevalence from a classifier's outputs is a task in its own right ---
quantification --- in which simple counting is known to be biased under
shift and dedicated estimators are used in its place (Esuli et al.
2023); the figure here is a proportion over a reviewed subset rather
than such an estimate, which is a further reason to carry it as
provisional.

Several limitations bound the reading, and they are stated in the
direction their optimism runs. The ground truth throughout is drawn from
a validator review queue that is itself organised by model confidence,
so both the incumbent's real operational figure and the sealed
operational-prior read rest on a review-selected pool rather than a
uniform sample of the stream; the selection runs optimistic in a
specific direction: because the queue preferentially surfaces the scenes
the model scores most confidently positive, the adjudicated pool is
enriched with the high-confidence flags on which precision is naturally
highest, while the lower-confidence flags that still clear the threshold
--- and fail more often --- are under-reviewed, so the operational
precisions quoted lean high against the full stream rather than
conservative. The general form is a risk estimate built on data the
model itself selected, for which inverse-acquisition-probability
weighting is unbiased provided the selection leaves non-zero probability
on every candidate (Farquhar et al. 2021); the queue selects
deterministically within a band, so no such probability exists, the
correction cannot be applied to a window already adjudicated, and the
remedy is prospective --- a softmax or epsilon-greedy acquisition, which
the service must adopt before a later window can be read this way. That
selection reaches the scoring prior itself: 0.05 is an estimate from the
adjudicated pool rather than a stream census, so cell (2) is composed at
a figure the service is still discovering, and the prior-sensitivity
band of Sect. 5.3 shows the scale of that dependence. The incumbent's
balanced cell carries a second caveat of its own: the membership of its
training set is not now recoverable, so its balanced-test figure is
reported as the deployed network's own evaluation and is not equated
with any headline from the predecessor paper, whose network was larger
and different. A third is conditional rather than fixed: the
recall-floor discipline tolerates missed positives because reprocessing
recovers them on a later pass, but that recovery holds only while the
model still generalises to sites it has not seen --- an overfit model
would keep missing novel-site positives, which reprocessing could then
never recover. The out-of-time read speaks to this directly: on
footprint-disjoint new sites the model's discrimination held (Sect.
5.3), evidence that it is not blind to unseen geography, though the
positive count there is too small to quantify new-site recall. Holding
capacity back (Sects 4.4, 3.2) is the design response to this risk
rather than a proof against it. And the loop is now closed without being
vindicated: the operational-prior figure predicted the fielded one in
direction but not in magnitude (above), so the commitment stands
discharged rather than confirmed.

The operating-point decay is a limitation of a different kind, because
it is a finding about deployment rather than about the evaluation. A
threshold fixed on the development set is not a set-and-forget quantity:
the out-of-time read shows its precision can fall while the model's
discrimination is intact, which means the operating point must be
validated against a current slice of the catalogue at deployment and
treated as something to recalibrate on a schedule, not inherited
unchanged from development. Prior correction and calibration are the
right tools for this: they govern the placement of the operating point,
which is distinct from the reported curve they cannot move (Sect. 4.3)
(Saerens et al. 2002; Heiser et al. 2020) --- effective, on systematic
reassessment, where the scheme has few classes and the classifier is
calibrated, and degrading sharply where it does not (Esuli et al. 2020).
This is the concrete operational requirement the paper hands to the
service, and it is the check the service must carry into its next
deployment.

The reporting method is not specific to internal waves, to Sentinel-1,
or even to a precision-first stance. Its one precondition is prior
shift: an operational classifier trained and reported under one class
balance but deployed against a materially different one. Wherever that
holds, the three-number axis applies as a portable reporting convention
--- report the balanced figure if convention demands it, but report
alongside it the operational-prior figure and commit to the
post-deployment number, so that a reader sees the prior effect rather
than having to infer it. The precision-first operating discipline of
this particular service (Sect. 2.2) is a separate matter --- the current
end of an inversion from the recall-first stance of the service's early
life, and one expected to normalise as the costs of the two errors
converge --- and the method does not depend on it: a service that
weighted recall, or weighted the two equally, would report the same
three numbers and read the same contrast. This echoes the
transferable-methodology framing of the predecessor paper, carried now
from pipeline selection to performance reporting.

Finally, the development cycle carries a geoscience point that outlives
the particular model. The decision not to spend capacity past the point
where it stopped paying was a scientific choice as much as a
regularisation one: a network overfit to the recurrence sites the
archive already knows would stop generalising to the unseen sites the
service exists to find, which is why the split holds sites out and why
the endpoint was matched to the data rather than maximised (Sects 4.4,
4.6). That the binding constraint ended the cycle back on the size of
the validated dataset, rather than on architecture, is itself the
finding that sets the next cycle's direction --- more validated data
from the growing archive, rather than a larger network on the data in
hand.

\subsection{7. Conclusions}\label{conclusions}

Balanced-test accuracy is the wrong number to trust for an operational
classifier whose deployment stream is skewed away from the prior it was
evaluated under. The proposal here is to stop reporting it alone:
characterise the model by three figures --- its balanced-test
performance, its performance on a frozen test set drawn at the
operational prior, and its post-deployment operational performance ---
and read the contribution in the contrast between them. The balanced
figure is exposed as optimistic, the operational-prior figure is built
to predict the fielded one, and the distance between the two is the
prior effect made visible rather than left to be inferred.

We demonstrated the instrument across a full precision-first development
cycle on the Internal Waves Service, reporting each lever for what it
did and did not improve. Widening negative variety, doubling width and
replacing the pooling head each lifted precision at the fixed recall
floor; shifting the training ratio toward the operational prior did not,
nor did width past the first doubling; and probability calibration could
not, by construction, being a monotone transform of the score at a
recall-pinned operating point. The negatives are as much a result as the
promotions --- they locate the mismatch as an evaluation problem rather
than a training one.

The figures are only as trustworthy as the discipline behind them. The
dataset was pinned before it was split, spatial leakage was controlled
at the true image footprint, the operating point was fixed on
development data, and the sealed evaluation set was read exactly once at
a pre-registered threshold, with every read's scoring script and test
committed before any number was seen. The out-of-time check applied that
same discipline, catching a real decay in the fixed operating point that
a balanced evaluation would not have surfaced.

Two things follow for the next cycle. The loop is now closed, and it
closed imperfectly: the promoted model's operational-prior figure of
0.927 anticipated a fielded precision it did not reach, the deployment
window returning 0.7936. The direction was right and the magnitude was
not, which is the instrument reporting on itself. And the binding
constraint has moved. Added capacity stopped clearing the promotion
margin and the aggregation lever is spent, yet no architectural ceiling
was found; what limits precision now is the size of the reducible
false-positive mass, which is to say the validated data. The next cycle
therefore restarts not from a larger network but from more of the
growing archive --- and, at deployment, from an operating point
revalidated against a current catalogue slice rather than inherited from
development. The operational prior is on the same footing: the rate the
instrument scores against is an estimate the service is still
discovering, so measuring it is work this method generates rather than
assumes.

\subsection{Back matter}\label{back-matter}

\textbf{Code and data availability.} The model and evaluation code, the
frozen train/development/lockbox split manifests, the per-experiment
generators and tests, and the results artefacts underlying all figures
and Table 1 are archived at Zenodo
(https://doi.org/10.5281/zenodo.21626953) and developed openly at
https://github.com/AIRCentre/iws-prior-matched-eval-paper under the
Apache-2.0 licence. The Sentinel-1 Wave-mode vignettes are Copernicus
data, freely available from ESA and referenced by acquisition identifier
in the manifests; they are not redistributed here. The sealed evaluation
lockbox is held under a single-read protocol and its imagery is not
published; its manifest and per-file checksums are included in the
archive, and the scores underlying the certification are provided.

\textbf{Author contributions.} JP: Conceptualization, Methodology,
Software, Formal analysis, Investigation, Visualization, Writing --
original draft, Supervision. JG: Software, Resources --- built and
operates the Internal Waves Service platform. AS: Resources --- provided
the initial classifier the work started from. ASF: Validation --- expert
adjudication of detections. All authors contributed to Writing -- review
and editing.

\textbf{Competing interests.} The authors declare that they have no
competing interests.

\textbf{Acknowledgements.} Not applicable.

\textbf{Use of AI tools.} AI-assisted tooling was used in developing the
software, including its test suite, and for paragraph-level rewriting
during manuscript preparation; all experimental design, scientific
decisions, and final content are the authors'.

\subsection*{References}\label{references}
\addcontentsline{toc}{subsection}{References}

\protect\phantomsection\label{refs}


\begin{CSLReferences}{1}{1}
\bibitem[\citeproctext]{ref-barintag2023}
Barintag, S., Z. An, Q. Jin, X. Chen, M. Gong, and T. Zeng. 2023.
{``{MTU2-Net}: Extracting Internal Solitary Waves from {SAR} Images.''}
\emph{Remote Sensing} 15 (23): 5441.
\url{https://doi.org/10.3390/rs15235441}.

\bibitem[\citeproctext]{ref-branco2016}
Branco, Paula, Luís Torgo, and Rita P. Ribeiro. 2016. {``A Survey of
Predictive Modeling on Imbalanced Domains.''} \emph{ACM Computing
Surveys} 49 (2). \url{https://doi.org/10.1145/2907070}.

\bibitem[\citeproctext]{ref-dockes2021}
Dockès, J., G. Varoquaux, and J.-B. Poline. 2021. {``Preventing Dataset
Shift from Breaking Machine-Learning Biomarkers.''} \emph{GigaScience}
10 (9): giab055. \url{https://doi.org/10.1093/gigascience/giab055}.

\bibitem[\citeproctext]{ref-esuli2023}
Esuli, Andrea, Alessandro Fabris, Alejandro Moreo, and Fabrizio
Sebastiani. 2023. \emph{Learning to Quantify}. Vol. 47. The Information
Retrieval Series. Springer, Cham.
\url{https://doi.org/10.1007/978-3-031-20467-8}.

\bibitem[\citeproctext]{ref-esuli2020}
Esuli, Andrea, Alessio Molinari, and Fabrizio Sebastiani. 2020. {``A
Critical Reassessment of the {Saerens-Latinne-Decaestecker} Algorithm
for Posterior Probability Adjustment.''} \emph{ACM Transactions on
Information Systems} 39 (2). \url{https://doi.org/10.1145/3433164}.

\bibitem[\citeproctext]{ref-farquhar2021}
Farquhar, Sebastian, Yarin Gal, and Tom Rainforth. 2021. {``On
Statistical Bias in Active Learning: How and When to Fix It.''}
\emph{International Conference on Learning Representations (ICLR)}.
\url{https://arxiv.org/abs/2101.11665}.

\bibitem[\citeproctext]{ref-heiser2020}
Heiser, T. J. T., M.-L. Allikivi, and M. Kull. 2020. {``Shift Happens:
Adjusting Classifiers.''} \emph{Machine Learning and Knowledge Discovery
in Databases (ECML PKDD 2019)}, Lecture notes in computer science, vol.
11907: 55--70. \url{https://doi.org/10.1007/978-3-030-46147-8_4}.

\bibitem[\citeproctext]{ref-kang2020}
Kang, B., S. Xie, M. Rohrbach, et al. 2020. {``Decoupling Representation
and Classifier for Long-Tailed Recognition.''} \emph{International
Conference on Learning Representations (ICLR)}.
\url{https://arxiv.org/abs/1910.09217}.

\bibitem[\citeproctext]{ref-kattenborn2022}
Kattenborn, T., F. Schiefer, J. Frey, H. Feilhauer, M. D. Mahecha, and
C. F. Dormann. 2022. {``Spatially Autocorrelated Training and Validation
Samples Inflate Performance Assessment of Convolutional Neural
Networks.''} \emph{ISPRS Open Journal of Photogrammetry and Remote
Sensing} 5: 100018. \url{https://doi.org/10.1016/j.ophoto.2022.100018}.

\bibitem[\citeproctext]{ref-maxwell2021a}
Maxwell, Aaron E., Timothy A. Warner, and Luis Andrés Guillén. 2021a.
{``Accuracy Assessment in Convolutional Neural Network-Based Deep
Learning Remote Sensing Studies---Part 1: Literature Review.''}
\emph{Remote Sensing} 13 (13): 2450.
\url{https://doi.org/10.3390/rs13132450}.

\bibitem[\citeproctext]{ref-maxwell2021b}
Maxwell, Aaron E., Timothy A. Warner, and Luis Andrés Guillén. 2021b.
{``Accuracy Assessment in Convolutional Neural Network-Based Deep
Learning Remote Sensing Studies---Part 2: Recommendations and Best
Practices.''} \emph{Remote Sensing} 13 (13): 2591.
\url{https://doi.org/10.3390/rs13132591}.

\bibitem[\citeproctext]{ref-pinelo2025}
Pinelo, J., A. M. Santos-Ferreira, J. Gonçalves, et al. 2025. {``{IWS}
--- {Internal Waves Service}: A World-First Repository for
Planetary-Scale Internal Solitary Waves Monitoring.''} \emph{Remote
Sensing for Agriculture, Ecosystems, and Hydrology XXVII}, Proc. SPIE,
vol. 13666: 1366605. \url{https://doi.org/10.1117/12.3069138}.

\bibitem[\citeproctext]{ref-pinelo2026}
Pinelo, J., A. Shukla, G. Titericz, A. Santos-Ferreira, J. Gonçalves,
and J. Moniz. 2026. {``Choosing an Operational Inference Pipeline for
Internal Solitary Wave Detection in {Sentinel-1} {SAR} Imagery:
{EVA02-Large}+{XGBoost} Versus {SAR\_CNN} V2 ({Lux.jl}).''}
\emph{EGUsphere {[}Preprint{]}}.

\bibitem[\citeproctext]{ref-roberts2017}
Roberts, D. R., V. Bahn, S. Ciuti, et al. 2017. {``Cross-Validation
Strategies for Data with Temporal, Spatial, Hierarchical, or
Phylogenetic Structure.''} \emph{Ecography} 40 (8): 913--29.
\url{https://doi.org/10.1111/ecog.02881}.

\bibitem[\citeproctext]{ref-saerens2002}
Saerens, M., P. Latinne, and C. Decaestecker. 2002. {``Adjusting the
Outputs of a Classifier to New a Priori Probabilities: A Simple
Procedure.''} \emph{Neural Computation} 14 (1): 21--41.
\url{https://doi.org/10.1162/089976602753284446}.

\bibitem[\citeproctext]{ref-saito2015}
Saito, Takaya, and Marc Rehmsmeier. 2015. {``The Precision-Recall Plot
Is More Informative Than the {ROC} Plot When Evaluating Binary
Classifiers on Imbalanced Datasets.''} \emph{PLOS ONE} 10 (3): e0118432.
\url{https://doi.org/10.1371/journal.pone.0118432}.

\bibitem[\citeproctext]{ref-santos-ferreira2025}
Santos-Ferreira, A. M., J. Pinelo, J. C. B. da Silva, et al. 2025.
{``The {Internal Waves Service} Workshop: Observing Internal Waves
Globally with Deep Learning and Synthetic Aperture Radar.''}
\emph{Bulletin of the American Meteorological Society}, E1462.
\url{https://doi.org/10.1175/BAMS-D-25-0133.1}.

\bibitem[\citeproctext]{ref-valavi-blockcv}
Valavi, R., J. Elith, J. J. Lahoz-Monfort, and G. Guillera-Arroita.
2019. {``{blockCV}: An {R} Package for Generating Spatially or
Environmentally Separated Folds for k-Fold Cross-Validation of Species
Distribution Models.''} \emph{Methods in Ecology and Evolution} 10 (2):
225--32. \url{https://doi.org/10.1111/2041-210X.13107}.

\end{CSLReferences}
\end{document}